# Graphical readings of possibilistic logic bases

**Salem Benferhat, Didier Dubois, Souhila Kaci and Henri Prade**
Institut de Recherche en Informatique de Toulouse (I.R.I.T.)–C.N.R.S.
Université Paul Sabatier, 118 route de Narbonne 31062 TOULOUSE Cedex 4, FRANCE
E-mail:{benferhat, dubois, kaci, prade}@irit.fr

## Abstract

Possibility theory offers either a qualitative, or a numerical framework for representing uncertainty, in terms of dual measures of possibility and necessity. This leads to the existence of two kinds of possibilistic causal graphs where the conditioning is either based on the minimum, or on the product operator. Benferhat et al. [3] have investigated the connections between min-based graphs and possibilistic logic bases (made of classical formulas weighted in terms of certainty). This paper deals with a more difficult issue: the product-based graphical representation of possibilistic bases, which provides an easy structural reading of possibilistic bases.

## 1 Introduction

Possibilistic logic offers a general framework for representing prioritized information by means of classical logical formulas which are associated with weights belonging to a linearly ordered scale (and which are handled according to the laws of possibility theory [21]). This leads to a stratification of the information base into layers of formulas according to the strength of the associated weights. This framework can be useful for representing knowledge, and then a weight represents the certainty level with which the associated formula is held for true. A possibilistic logic base can also represent desires or goals having different levels of priority.

A set of weighted logical formulas, constituting a possibilistic logic base is semantically equivalent to a possibility distribution which rank-orders the possible worlds according to their levels of possibility. These possibility levels are to be understood as plausibility or normality levels in case the base gathers pieces of knowledge, or as levels of satisfaction of reaching a considered world in case it is a collection of weighted goals.

Apart from the syntactic representation provided by a possibilistic base, an other compact representation, sharing the same semantics is of interest, namely, possibilistic directed acyclic graphs (DAGs) [12, 3]. The merit of the DAG representation is, as usual, to exhibit some independence structure (here in the possibilistic framework), and to provide a structured decomposition of the possibility distribution underlying the base. However, the interest of working with different representation modes has been pointed out in several works [4, 16].

Depending if we are using a numerical scale such as [0,1], or a simple linearly ordered scale, two types of conditioning can be defined in possibility theory; one based on the product which requires a numerical scale, and one based on minimum operation for which any linearly ordered scale fits. This corresponds to quantitative and qualitative possibility theory respectively [9]. In quantitative possibility theory, possibility degrees can be viewed as upper bounds of probabilities [8]. Until recently, quantitative possibility theory did not have any operational semantics strictly speaking, despite an early proposal by Giles [13] in the setting of upper and lower probabilities, recently taken over by Walley, De Cooman and Aeyels [20, 5]. One way to avoid the measurement problem is to develop a qualitative epistemic possibility theory where only ordering relations are used [9]. For quantitative (subjective) possibilities, an operational semantics has been recently proposed [18], [10] which differs from the upper and lower probabilistic setting proposed by Giles and followers. It is based on the semantics of the transferable belief model [17], itself based on betting odds. It can be shown that the least informative among the belief structures that are compatible with prescribed betting rates is a possibility measure. Then, it can be also proved that the min-based idempotent conjunctive combination of two possibility measures cor-responds to the hyper-cautious conjunctive combination of the belief functions induced by the possibility measures.

The translation of a possibilistic graph (product-based, or minimum-based) into a possibilistic logic base has been already provided [2], as well as the converse transformation for minimum-based possibilistic graphs [3]. The translation of a possibilistic logic base into a product-based graph, which is less straightforward is now addressed in this paper. The transformation is illustrated on a running example dealing with the goals of an agent.

The paper is organized as follows. After a minimal background on possibility theory, possibilistic logic and possibilistic graphs, the transformation of a possibilistic logic base into a product-based possibilistic graph is discussed in details in the rest of the paper.

## 2 Background

### 2.1 Possibility theory

Let $\mathcal{L}$ be a finite propositionnal language. $\Omega$ is the set of all classical interpretations. Greek letters $\phi, \psi \cdots$ denote formulas. The notation $\omega \models \phi$ means that $\omega$ is a model of $\phi$.

A possibility distribution [21] $\pi$ is a function mapping a set of interpretations (or worlds) $\Omega$ into a linearly ordered scale, usually the interval $[0, 1]$. $\pi(\omega)$ represents the degree of compatibility of the interpretation $\omega$ with the available beliefs about the real world in a case of uncertain information, or the satisfaction degree of reaching a state $\omega$, when modelling preferences. $\pi(\omega) = 1$ means that it is totally possible for $\omega$ to be the real world (or that $\omega$ is fully satisfactory), $1 > \pi(\omega) > 0$ means that $\omega$ is only somewhat possible (or satisfactory), while $\pi(\omega) = 0$ means that $\omega$ is certainly not the real world (or not satisfactory at all). A possibility distribution is said to be normalized (or consistent) if $\exists \omega$ s.t. $\pi(\omega) = 1$. Only normalized distributions are considered here.

Given a possibility distribution $\pi$, two dual measures are defined which rank order the formulas of the language:

1. The possibility (or consistency) measure of a formula $\phi$:
$$\Pi(\phi) = max\{\pi(\omega) : \omega \models \phi\},$$
which evaluates the extent to which $\phi$ is consistent with the available beliefs expressed by $\pi$.

2. The necessity (or certainty) measure of a formula $\phi$:
$$N(\phi) = 1 - \Pi(\neg \phi),$$
which evaluates the extent to which $\phi$ is entailed by the available beliefs.

When dealing with knowledge, a statement $\phi$ is thus estimated in terms of two measures $\Pi$ and $N$ which enable us to differenciate between the certainty of $\neg \phi$ ($N(\neg \phi) = 1$) and the total lack of certainty in $\phi$ ($N(\phi) = 0$). When dealing with desires $N(\phi)$ refers to the imperativeness of goals $\phi$, while $\Pi(\phi)$ estimates how satisfactory is to reach $\phi$.

The definition of conditioning in possibility theory depends if we use an ordinal, or a numerical scale. In an ordinal setting, *min-based conditioning* is used and is defined as follows:

$$\Pi(\psi \mid \phi) = \begin{cases} 1 & \text{if } \Pi(\phi \wedge \psi) = \Pi(\phi) \\ \Pi(\phi \wedge \psi) & \text{if } \Pi(\phi \wedge \psi) < \Pi(\phi) \end{cases}$$

In a numerical setting, the *product-based conditioning* is used:
$$\Pi(\psi \mid_\times \phi) = \frac{\Pi(\phi \wedge \psi)}{\Pi(\phi)}.$$

Moreover, if $\Pi(\phi) = 0$, then $\Pi(\psi \mid \phi) = \Pi(\neg \psi \mid \phi) = 1$.

Both forms of conditioning satisfy an equation of the form: $\Pi(\psi) = \Box(\Pi(\psi \mid \phi), \Pi(\phi))$, which is similar to Bayesian conditioning, for $\Box = min$ or $product$. In this paper we privilege the numerical setting.

It is clear that $\pi'(\omega) = \pi(\omega \mid \phi)$, the result of conditioning a possibility distribution $\pi$ with $\phi$ is always normalized.

### 2.2 Possibilistic knowledge bases

A possibilistic knowledge base is a set of weighted formulas of the form $\Sigma = \{(\phi_i, \alpha_i) : i = 1, n\}$ where $\phi_i$ is a classical formula and, $\alpha_i$ belongs to $[0, 1]$ in a numerical setting and represents the level of certainty or priority attached to $\phi_i$.

Given a possibilistic base $\Sigma$, we can generate a unique possibility distribution $\pi_\Sigma$, where interpretations will be ranked w.r.t. the highest formula that they falsify, namely [6]:

**Definition 1** $\forall \omega \in \Omega$,
$$\pi_\Sigma(\omega) = \begin{cases} 1 & \text{if } \forall (\phi_i, \alpha_i) \in \Sigma, \omega \models \phi_i \\ 1 - max\{\alpha_i : (\phi_i, \alpha_i) \in \Sigma \text{ and } \omega \not\models \phi_i\} & \text{otherwise.} \end{cases}$$

**Example 1** *Let $su, wi, se$ be three symbols which stand for "sun", "wind" and "sea" respectively. Let $\Sigma$ be the following possibilistic base:*

$\Sigma = \{(su \vee \neg wi, \frac{2}{3}), (\neg wi \vee se, \frac{1}{3}), (wi \vee \neg se, \frac{1}{3}), (su \vee se, \frac{1}{3})\}.$

*These rules express the goals of somebody who likes basking in the sun, or going windsurfing (which requires wind and sea). The most prioritized formula expresses that the person strongly dislikes to have wind in a non-suny day, while the other less prioritary goals express that she dislikes situations with wind but without sea, or with sea without wind, or with neither sun nor sea.*

Let $\Omega = \{\omega_0 = su \wedge \neg wi \wedge \neg se, \omega_1 = su \wedge \neg wi \wedge se, \omega_2 = su \wedge wi \wedge \neg se, \omega_3 = su \wedge wi \wedge se, \omega_4 = \neg su \wedge \neg wi \wedge \neg se, \omega_5 = \neg su \wedge \neg wi \wedge se, \omega_6 = \neg su \wedge wi \wedge \neg se, \omega_7 = \neg su \wedge wi \wedge se\}$.
Let $\pi_\Sigma$ be the possibility distribution associated with $\Sigma$: $\pi_\Sigma(\omega_0) = \pi_\Sigma(\omega_3) = 1$, $\pi_\Sigma(\omega_1) = \pi_\Sigma(\omega_2) = \pi_\Sigma(\omega_4) = \pi_\Sigma(\omega_5) = \frac{2}{3}$ and $\pi_\Sigma(\omega_6) = \pi_\Sigma(\omega_7) = \frac{1}{3}$.

The converse transformation from $\pi$ to $\Sigma$ is straightforward. Let $1 > \beta_1 > \cdots > \beta_n \geq 0$ be the different weights used in $\pi$. Let $\phi_i$ be a classical formula whose models are those having the weight $\beta_i$ in $\pi$. Let $\Sigma = \{(\neg\phi_i, 1 - \beta_i) : i = 1, n\}$. Then, $\pi_\Sigma = \pi$.

We now give further definitions which will be used latter in the paper:

**Definition 2** *Let $\Sigma$ be a possibilistic knowledge base, and $\alpha \in [0,1]$. We call the $\alpha$-cut (resp. strict $\alpha$-cut) of $\Sigma$, denoted by $\Sigma_{\geq \alpha}$ (resp. $\Sigma_{>\alpha}$), the set of classical formulas in $\Sigma$ having a certainty degree at least equal to $\alpha$ (resp. strictly greater than $\alpha$).*

$Inc(\Sigma) = max\{\alpha_i : \Sigma_{\geq \alpha_i} \text{ is inconsistent}\}$ denotes the inconsistency degree of $\Sigma$. When $\Sigma$ is consistent, we have $Inc(\Sigma) = 0$.
Subsumption can now be defined:

**Definition 3** *Let $(\phi, \alpha)$ be a formula in $\Sigma$. Then, $(\phi, \alpha)$ is said to be subsumed in $\Sigma$ if $(\Sigma - \{(\phi, \alpha)\})_{\geq \alpha} \vdash \phi$. $(\phi, \alpha)$ is said to be strictly subsumed in $\Sigma$ if $\Sigma_{>\alpha} \vdash \phi$.*

Indeed, we have the following proposition [6]:

**Proposition 1** *Let $\Sigma$ be a possibilistic knowledge base, and $(\phi, \alpha)$ be a subsumed formula in $\Sigma$. Let $\Sigma' = \Sigma - \{(\phi, \alpha)\}$. Then, $\Sigma$ and $\Sigma'$ are equivalent i.e. $\pi_\Sigma = \pi_{\Sigma'}$.*

### 2.3 Possibilistic networks

Another possibilistic representation framework of uncertain information is graphical and is based on conditioning. Information is then represented by possibilistic DAGs (Directed Acyclic Graphs) [3, 12], where nodes represent variables (in this paper, we assume that they are binary), and edges express influence links between variables. When there exists a link from $A$ to $B$, $A$ is said to be a parent of $B$. The set of parents of a given node $A$ is denoted by $Par(A)$. By the capital letter $A$ we denote a variable which represents either the symbol $a$ or its negation. An interpretation in this section will be simply denoted by $A_1 \cdots A_n$, or by $\omega$.

Uncertainty is expressed at each node in the following way:
- For root nodes $A_i$ (namely $Par(A_i) = \emptyset$) we provide the prior possibility of $a_i$ and of its negation $\neg a_i$. These priors should satisfy the normalization condition:

$$max(\Pi(a_i), \Pi(\neg a_i)) = 1.$$

- For other nodes $A_j$, we provide conditional possibility of $a_j$ and of its negation $\neg a_j$ given any complete instantiation of each variable of parents of $A_j$, $\omega_{Par(A_j)}$. These conditional possibilities should also satisfy the normalization condition:

$$max(\Pi(a_j \mid \omega_{Par(A_j)}), \Pi(\neg a_j \mid \omega_{Par(A_j)})) = 1.$$

A product-based possibilistic graph, denoted by $\Pi G$, induces a unique joint distribution using a so-called *chain rule*:

**Definition 4** *Let $\Pi G$ be a product-based possibilistic graph. The joint possibility distribution associated with $\Pi G$ is computed with the following equation (called chain rule):*
$\pi(\omega) = *\{\Pi(a \mid \omega_{Par(A)}) : \omega \models a \text{ and } \omega \models \omega_{Par(A)}\}$,
*where $* = product$.*

The converse transformation from a possibility distribution $\pi$ to a product-based graph is straightforward. Indeed, given an interpretation $\omega = A_1 A_2 \cdots A_n$, we have:

$$\pi(A_1 \cdots A_n) = \pi(A_1 \mid A_2 A_3 \cdots A_n) * \pi(A_2 A_3 \cdots A_n).$$

Applying repeatedly the Bayesian-like rule for an arbitrarily ordering $A_1, \cdots, A_n$ between variables, we get:
$\pi(A_1 \cdots A_n) = \pi(A_1 \mid A_2 \cdots A_n) *$
$\pi(A_2 \mid A_3 \cdots A_n) * \cdots * \pi(A_{n-1} \mid A_n) * \pi(A_n)$.

This decomposition is possible since the *product* is associative. The decomposition leads to a causal possibilistic graph where parent of $A_i$ are $\{A_{i+1}, \cdots, A_n\}$. It is clear that for the same joint distribution, we can construct several causal possibilistic graphs depending on the choice of the ordering between variables.
In possibility theory, there are several definitions of independence relations (see e.g. [11],[1]). In this paper, since we deal with product-based conditioning, we will use the following definition of independence, namely: $\phi$ and $\psi$ are independent if $\Pi(\psi \mid \phi) = \Pi(\psi)$.

## 3   From possibilistic bases to product-based graphs : Basic ideas

In [2], the authors have provided the transformation of possibilistic graphs to possibilistic bases. In [3], the transformation from $\Sigma$ to a min-based graph, where conditioning is based on minimum operation rather than on the product, has been given.

In the following, we provide the transformation from a possibilistic base $\Sigma$ to a product-based graph $\Pi G$. This transformation is different from the transformation in case of conditioning based on the minimum operation. However there is one common step in both approaches which consists in putting the knowledge base into a clausal form and in removing tautologies.
The following proposition shows how to put the base in a clausal form:

**Proposition 2** *[6] Let $\Sigma$ be a possibilistic base. Let $(\phi, \alpha)$ be a formula in $\Sigma$, and $\{\phi_1, \cdots, \phi_n\}$ be the set of clauses encoding $\phi$. We define $\Sigma'$ obtained by replacing each formula $(\phi, \alpha)$ in $\Sigma$ by $\{(\phi_1, \alpha), \cdots, (\phi_n, \alpha)\}$. Then, $\Sigma$ and $\Sigma'$ are equivalent i.e., $\pi_\Sigma = \pi_{\Sigma'}$.*

The removing of tautologies is important since it avoids fictitious dependence relations between variables. For example, the tautological formula $(\neg x \vee \neg y \vee x, 1)$ might induce a link between $X$ and $Y$.

The basic idea in the transformation from the knowledge base to a possibilistic graph is first to fix an arbitrarily ordering of variables $A_1, \cdots, A_n$. This ordering means that the parents of $A_i$ should be among $A_{i+1}, \cdots, A_n$ (however it can be empty). Then we proceed by successive decompositions of $\Sigma$, which is associated with the above decomposition of a possibility distribution $\pi$, namely:

$$\pi(A_1 \cdots A_n) = \pi(A_1 \mid A_2 \cdots A_n) * \pi(A_2 \cdots A_n).$$

The result of each step $i$ is a knowledge base associated with $\pi(A_i \cdots A_n)$. Moreover, this resulting base will enable us to determine $Par(A_i)$ the set of parents of $A_i$ and to compute the conditional possibilities $\pi(A_i \mid Par(A_i))$.

In the following, we will only consider the first step ($i = 1$), and we will denote by $\Sigma_C$ (C for Current) the base associated with $\pi(A_2 \cdots A_n)$. The procedure of this first step can be then iterated at each step. $\Sigma_C$ is called the marginalized base.
The computation of $\Sigma_C$ is provided in the next section, then in Section 5 we show how to determine parents of each node, and lastly in Section 6 we compute the conditional possibility degrees.

## 4   Computation of the marginalized base $\Sigma_C$

The computation of $\Sigma_C$ involves three tasks:

- decomposing a possibility distributions $\pi$ into its restriction on $a_1$ and its negation $\neg a_1$ (recall that $a_1$ and $\neg a_1$ are the two only instances of $A_1$),
- marginalisation of these two distributions for getting rid of $A_1$,
- the effective computation of $\Sigma_C$.

### 4.1   Decomposing $\pi$

Let us first define two possibility distributions $\pi_{a_1}$ and $\pi_{\neg a_1}$ in the following way:

$$\pi_{a_1}(\omega) = \begin{cases} \pi(\omega) & \text{if } \omega \models a_1 \\ 0 & otherwise. \end{cases}$$

$$\pi_{\neg a_1}(\omega) = \begin{cases} \pi(\omega) & \text{if } \omega \models \neg a_1 \\ 0 & otherwise. \end{cases}$$

$\pi_{a_1}$ (resp. $\pi_{\neg a_1}$) are very similar to conditioning except that we do not normalize after learning $a_1$ (resp. $\neg a_1$). $\pi_{a_1}$ and $\pi_{\neg a_1}$ are simply the decomposition of $\pi$. Indeed, it can be checked that:

$$\pi(\omega) = max(\pi_{a_1}(\omega), \pi_{\neg a_1}(\omega)), \forall \omega.$$

**Example 2** *We consider again the possibilistic base $\Sigma$ of Example 1. Let us compute $\pi_{se}$ and $\pi_{\neg se}$. We get:*
*$\pi_{se}(\omega_0) = \pi_{se}(\omega_2) = \pi_{se}(\omega_4) = \pi_{se}(\omega_6) = 0;$*
*$\pi_{se}(\omega_7) = \frac{1}{3}; \pi_{se}(\omega_1) = \pi_{se}(\omega_5) = \frac{2}{3}$ and $\pi_{se}(\omega_3) = 1$.*
*Also,*
*$\pi_{\neg se}(\omega_1) = \pi_{\neg se}(\omega_3) = \pi_{\neg se}(\omega_5) = \pi_{\neg se}(\omega_7) = 0;$ $\pi_{\neg se}(\omega_6) = \frac{1}{3}; \pi_{\neg se}(\omega_2) = \pi_{\neg se}(\omega_4) = \frac{2}{3}$ and $\pi_{\neg se}(\omega_0) = 1.$*
*Then, we can check that*
*$\forall \omega, \pi_\Sigma(\omega) = max(\pi_{se}(\omega), \pi_{\neg se}(\omega))$ where $\pi_\Sigma$ is the possibility distribution associated with $\Sigma$.*

We can easily check that the possibilitic bases associated with these distributions are given by:

**Proposition 3** *The possibilistic base associated with $\pi_{a_1}$ (resp. $\pi_{\neg a_1}$) is $\Sigma \cup \{(a_1, 1)\}$ (resp. $\Sigma \cup \{(\neg a_1, 1)\}$).*

**Proof**
The proof is obvious. Indeed, we have two cases:

- $\omega \not\models a_1$, then using Definition 1:
  $\pi_{\Sigma \cup \{(a_1, 1)\}}(\omega)$
  $= 1 - max\{a_i : (\phi_i, a_i) \in \Sigma \cup \{(a_1, 1)\}, \omega \not\models \phi_i\}$
  $= 0$ (since $\omega \not\models a_1$)

- $\omega \models a_1$, then using Definition 1:
$\pi_{\Sigma \cup \{(a_1,1)\}}(\omega) = 1 - max\{a_i : (\phi_i, a_i) \in \Sigma \cup \{(a_1,1)\}, \omega \not\models \phi_i\}$
$= 1 - max\{a_i : (\phi_i, a_i) \in \Sigma, \omega \not\models \phi_i\}$
$(since\ \omega \models a_1)$
$= \pi_\Sigma(\omega)$ □

The two following lemmas give two simplifications of $\Sigma \cup \{(a_1, 1)\}$ (resp. $\Sigma \cup \{(\neg a_1, 1)\}$).

**Lemma 1** *Let $\Sigma'_1 = \Sigma \cup \{(a_1, 1)\}$. Let $\Sigma''_1 = \Sigma'_1 - \{(a_1 \vee x, \alpha) : (a \vee x, \alpha) \in \Sigma\}$ a base obtained from $\Sigma'_1$ by removing clauses containing $a_1$. Then, $\Sigma'_1$ and $\Sigma''_1$ are equivalent, in the sense that they generate the same possibility distribution which is $\pi_{a_1}$.*

The proof is obvious since $(a_1 \vee x, \alpha)$ is subsumed by $(a_1, 1)$.

**Lemma 2** *Let $\Sigma'_1 = \Sigma \cup \{(a_1, 1)\}$. Let $\Sigma''_1$ be the possibilistic base obtained from $\Sigma'_1$ by replacing each clause of the form $(\neg a_1 \vee x, \alpha)$ by $(x, \alpha)$. Then, $\Sigma'_1$ and $\Sigma''_1$ are equivalent.*

The proof can be again easily checked. Indeed, $(a_1, 1)$ and $(\neg a_1 \vee x, \alpha)$ implies $(x, \alpha)$ which can be added to $\Sigma'_1$. And once $(x, \alpha)$ is added, $(\neg a_1 \vee x, \alpha)$ can be retrieved since it is subsumed by $(x, \alpha)$.

### 4.2 Marginalisation

In this section, we are interested in computing the marginal distributions from $\pi_{a_1}$ (resp. $\pi_{\neg a_1}$) defined on $\{A_2, \cdots, A_n\}$.
Let us denote $\Sigma_{a_1}$ (resp. $\Sigma_{\neg a_1}$) the result of the application of Lemmas 1 and 2 to $\Sigma \cup \{(a_1, 1)\}$ (resp. $\Sigma \cup \{(\neg a_1, 1)\}$), namely the result of removing clauses of the form $(x \vee a_1, \alpha)$ from $\Sigma$ and the result of replacing $(x \vee \neg a_1, \alpha)$ by $(x, \alpha)$. Therefore, the only clause in $\Sigma_{a_1}$ which contains $a_1$ is $(a_1, 1)$.
Let us denote $\pi^{A_1}_{a_1}, \pi^{A_1}_{\neg a_1}$ the result of marginalisation of $\pi_{a_1}$ and $\pi_{\neg a_1}$ on $\{A_2, \cdots, A_n\}$, namely:
$\pi^{A_1}_{a_1}(A_2 \cdots A_n) = \Pi_{a_1}(A_2 \cdots A_n)$
(resp. $\pi^{A_1}_{\neg a_1}(A_2 \cdots A_n) = \Pi_{\neg a_1}(A_2 \cdots A_n)$).

**Example 3** *Let $A_1 = SE$. Then,*
$\pi^{SE}_{se}(\neg su \wedge wi) = \frac{1}{3}$, $\pi^{SE}_{se}(su \wedge \neg wi) = \pi^{SE}_{se}(\neg su \wedge \neg wi) = \frac{2}{3}$, and $\pi^{SE}_{se}(su \wedge wi) = 1$.
*Also,*
$\pi^{SE}_{\neg se}(\neg su \wedge wi) = \frac{1}{3}$, $\pi^{SE}_{\neg se}(su \wedge wi) = \pi^{SE}_{\neg se}(\neg su \wedge \neg wi) = \frac{2}{3}$ and $\pi^{SE}_{\neg se}(su \wedge \neg wi) = 1$.

NB. $\Pi_{a_1}(A_2 \cdots A_n)$ is the possibility measure associated with $\pi_{a_1}$ defined on $\{A_1, \cdots, A_n\}$.

The following lemma provides the syntactic counterpart of $\pi^{A_1}_{a_1}$ (resp. $\pi^{A_1}_{\neg a_1}$):

**Lemma 3** *The possibilistic base associated with $\pi^{A_1}_{a_1}$ (resp. $\pi^{A_1}_{\neg a_1}$) is $\Sigma_{a_1} - \{(a_1, 1)\}$ (resp. $\Sigma_{\neg a_1} - \{(\neg a_1, 1)\}$).*

**Proof**
The proof is obvious. First, note that the only clause containing $a_1$ in $\Sigma_{a_1}$ is $(a_1, 1)$. Then,
$\pi^{A_1}_{a_1}(A_2 \cdots A_n) =$
$min\{1 - \alpha_i : (\phi_i, \alpha_i) \in \Sigma_{a_1} - \{(a_1, 1)\}, A_2 \cdots A_n \not\models \phi_i\}$
$= min\{1 - \alpha_i : (\phi_i, \alpha_i) \in \Sigma_{a_1} - \{(a_1, 1)\},$
$\quad a_1 A_2 \cdots A_n \not\models \phi_i\}$
$= max\{min\{1 - \alpha_i : (\phi_i, \alpha_i) \in \Sigma_{a_1},$
$\quad a_1 A_2 \cdots A_n \not\models \phi_i\},$
$\quad min\{1 - \alpha_i : (\phi_i, \alpha_i) \in \Sigma_{a_1},$
$\quad \neg a_1 A_2 \cdots A_n \not\models \phi_i\}\}$
$(since\ min\{1 - \alpha_i : (\phi_i, \alpha_i) \in \Sigma_{a_1},$
$\quad \neg a_1 A_2 \cdots A_n \not\models \phi_i\} = 0)$
$= max\{\pi_{a_1}(a_1 A_2 \cdots A_n), \pi_{a_1}(\neg a_1 A_2 \cdots A_n)\}$
$= \pi_{a_1}(A_2 \cdots A_n)$. □

### 4.3 Effective computation of $\Sigma_C$

Given Lemma 3 we are now able to provide the possibilistic base associated with $\pi(A_2 \cdots A_n)$ by noticing that:

$\pi(A_2 \cdots A_n) = max(\pi^{A_1}_{a_1}(A_2 \cdots A_n), \pi^{A_1}_{\neg a_1}(A_2 \cdots A_n))$.

**Example 4** *We again consider the possibility distributions $\pi^{SE}_{se}$ and $\pi^{SE}_{\neg se}$ computed in Example 3. We have:*
$\pi(su \wedge \neg wi) = \pi(su \wedge wi) = 1$, $\pi(\neg su \wedge \neg wi) = \frac{2}{3}$ and $\pi(\neg su \wedge wi) = \frac{1}{3}$.

**Proposition 4** *Let $\Sigma_1 = \Sigma_{a_1} - \{(a_1, 1)\}$ and $\Sigma_2 = \Sigma_{\neg a_1} - \{(\neg a_1, 1)\}$. The possibilistic base associated with $\pi(A_2, \cdots, A_n)$ is:*
$\Sigma_C = \{(\phi_i \vee \psi_j, min(\alpha_i, \beta_j) :$
$\quad (\phi_i, \alpha_i) \in \Sigma_1, (\psi_j, \beta_j) \in \Sigma_2)\}$.

**Proof**
The proof is obvious, indeed:
$\pi_{\Sigma_C}(\omega) = 1 - max\{min(\alpha_i, \beta_j) : (\phi_i, \alpha_i) \in \Sigma_1, (\psi_j, \beta_j) \in \Sigma_2, \omega \not\models \phi_i \vee \psi_j\}$
$= 1 - min\{max\{\alpha_i : (\phi_i, \alpha_i) \in \Sigma_1, \omega \not\models \phi_i\},$
$\quad max\{\beta_j : (\psi_j, \beta_j) \in \Sigma_2, \omega \not\models \psi_j\}\}$
$= max\{1 - max\{\alpha_i : (\phi_i, \alpha_i) \in \Sigma_1, \omega \not\models \phi_i\},$
$\quad 1 - max\{\beta_j : (\psi_j, \beta_j) \in \Sigma_2, \omega \not\models \psi_j\}\}$
$= max(\pi^{A_1}_{a_1}(\omega), \pi^{A_1}_{\neg a_1}(\omega))$. □

### 4.4 Summary

Let us now summarize the computation of $\Sigma_C$:

1. Add $(a_1, 1)$ (resp. $(\neg a_1, 1)$) to $\Sigma$,

2. Remove clauses containing $a_1$ (resp. $\neg a_1$) of the form $(a_1 \vee x, \alpha)$ (resp. $(\neg a_1 \vee x, \alpha)$),

3. Replace clauses of the form $(\neg a_1 \vee x, \alpha)$ (resp. $(a_1 \vee x, \alpha)$) by $(x, \alpha)$.

Let $\Sigma_1$ (resp. $\Sigma_2$) the result of the Step 3. Then,
$\Sigma_C = \{(\phi \vee \psi, min(\alpha, \beta)) : (\phi, \alpha) \in \Sigma_1 - \{(a_1, 1)\},$
$(\psi, \beta) \in \Sigma_2 - \{(\neg a_1, 1)\}\}$.

**Example 5** *Let us consider our example again.*
*Let* $\{SE, WI, SU\}$ *be the ordering of the variables (namely* $A_1 = SE$, $A_2 = WI$, *and* $A_3 = SU$*). We start with the variable* $SE$.
*Let us compute* $\Sigma_C$ *the possibilistic base associated with* $\pi_\Sigma(A_2 A_3)$ *(namely,* $\pi_\Sigma(WI \wedge SU)$*).*
*We first add "se" with a degree 1, we get:*

$\{(wi \vee \neg se, \frac{1}{3}), (\neg wi \vee se, \frac{1}{3}), (su \vee se, \frac{1}{3}), (su \vee \neg wi, \frac{2}{3}), (se, 1)\}$.
*Then we remove clauses containing se (except the added one), we get*
$\{(wi \vee \neg se, \frac{1}{3}), (su \vee \neg wi, \frac{2}{3}), (se, 1)\}$.
*Then we replace all clauses of the form* $(\phi \vee \neg se, \alpha)$ *by* $(\phi, \alpha)$ *we get:*
$\Sigma_{se} = \{(wi, \frac{1}{3}), (su \vee \neg wi, \frac{2}{3}), (se, 1)\}$.
*Similarly, for* $\neg se$*:*
$\Sigma_{\neg se} = \{(\neg wi, \frac{1}{3}), (su, \frac{1}{3}), (su \vee \neg wi, \frac{2}{3}), (\neg se, 1)\}$.
*Finally, using Proposition 3 we have:*

$\Sigma_1 = \{(wi, \frac{1}{3}), (su \vee \neg wi, \frac{2}{3})\}$,
$\Sigma_2 = \{(\neg wi, \frac{1}{3}), (su, \frac{1}{3}), (su \vee \neg wi, \frac{2}{3})\}$ *and*
$\Sigma_C = \{(su, \frac{1}{3}), (su \vee \neg wi, \frac{2}{3})\}$.
*It can be checked that* $\pi_{\Sigma_C}(WI \wedge SU) = \pi(WI \wedge SU)$, *where* $\pi$ *is the possibility distribution computed in Example 4.*

The problem of finding a syntactic counterpart of the marginalization process is clearly similar to the marginalization problem addressed in [15]. Their approach is more based on successive resolution in order to compute the marginalization base. Another small difference is that their approach is proposed in the classical case, without levels of priorities. Both procedures are polynomial.

## 5  Determining parents of $A_1$

We are interested in determining $Par(A_1)$ which are the parents of $A_1$. This set should be such that:
$\Pi(A_1 \mid A_2 \cdots A_n) = \Pi(A_1 \mid Par(A_1))$.
Once parents of $A_1$ are determined, we compute the conditional possibility degrees $\Pi(A_1 \mid Par(A_1))$ in Section 6.
The determination of parents of $A_1$ is done in an incremental way. First, we take $Par(A_1)$ as the set of variables which are directly involved at least in one clause containing $a_1$ or $\neg a_1$. $Par(A_1)$ are obvious parents of $A_1$. However, it may exist other "hidden" variables, whose observation influences the conditional possibility $\Pi(A_1 \mid Par(A_1))$. To show this, let us give the following illustration:

**Example 6** *Let* $\Sigma = \{(a_2 \vee a_1, .4), (a_3, .7)\}$, *then we can easily check in context* $\neg a_2$, $a_1$ *is deduced to a degree .4, which means that* $\Pi(\neg a_1 \mid \neg a_2) = 1 - .4 = .6$. *However, in context* $\neg a_2 \neg a_3$ *the certainty of* $a_1$ *is now 0, which means that* $\Pi(\neg a_1 \mid \neg a_2 \neg a_3) = 1 - 0 = 1$ *(due to presence of a conflict in* $\Sigma \cup \{\neg a_2 \wedge \neg a_3\}$*). Hence* $A_3$ *should also be considered as a parent of* $A_1$, *even if it is not directly involved. (see [6] for the presentation of the inference machinery in possibilistic logic).*

Algorithm 1 provides the computation of $Par(A_1)$.

**Algorithm 1:** Determining_Parents_of_$A_1$

Data: $\Sigma$
Result: Parents of $A_1$
**begin**
　**1.** Finding immediate parents of $A_1$
　　**1.a.** $\Sigma_{A_1} \leftarrow \{(\phi_i, \alpha_i) : (\phi_i, \alpha_i) \in \Sigma$
　　　　and $\phi_i$ contains either $a_1$ or $\neg a_1\}$.
　　**1.b.** $Par(A_1) \leftarrow \{V : \exists (\phi_i, \alpha_i) \in \Sigma_{A_1}$
　　　　containing an instance of $V\}$

　**2.** Checking for hidden parents
　　**2.a.** $B \leftarrow \Sigma$, let $(x_1, \cdots, x_n)$ be an instance of $Par(A_1)$
　　**2.b.** Remove from $B$ each clause containing $x_i$.
　　**2.c.** Replace in $B$ each clause of the form
　　　　$(\neg x_i \vee \neg x_j \cdots \vee \neg x_k \vee \phi, \alpha)$ by $(\phi, \alpha)$
　　**2.d.** Let $\alpha$ (resp. $\beta$) be the certainty degree of
　　　　$a_1$ (resp. $\neg a_1$) from $B \cup \{(x_1, 1), \cdots, (x_n, 1)\}$
　　**2.e.** if there exists $(\phi, \gamma) \in B$ such that $\gamma \geq \alpha$
　　　　(resp. $\gamma \geq \beta$), then:
　　　　$Par(A_1) = Par(A_1) \cup$
　　　　　　$\{V : \phi$ contains an instance of $V\}$
　Go to Step 2
　**return** $Par(A_1)$
**end**

Let us briefly explain this algorithm. The first step simply starts with parents of $A_1$, the set of variables which are directly linked with $A_1$. Step **2** checks for hidden parents of $A_1$, namely if $Par(A_1)$ can be extended or not.
A set of variables $V$ has to be added to $Par(A_1)$, if there exists an instance $(x_1, \cdots, x_n)$ for $Par(A_1)$ such that:
$\Pi(a_1 \mid x_1 x_2 \cdots x_n) \neq \Pi(a_1 \mid x_1 x_2 \cdots x_n v)$.
(resp. $\Pi(\neg a_1 \mid x_1 x_2 \cdots x_n) \neq \Pi(\neg a_1 \mid x_1 x_2 \cdots x_n v)$),
where $v$ is an instance of $V$. The computation of $\Pi(\neg a_1 \mid x_1 \cdots x_n)$ can be done syntactically from $\Sigma \cup \{(x_1, 1), \cdots, (x_n, 1)\}$ (for a formal computation of $\Pi(\neg a_1 \mid x_1 \cdots x_n)$ see Section 6).

This is what it is done in Step **2**, by checking if any additional variables can have influences on $\Pi(a_1 \mid x_1 \cdots x_n)$. To achieve this goal, we first assume that $(x_1, 1), \cdots, (x_n, 1)$ are true. Step **2.b** removes $(x_i \vee \phi, \alpha)$ since the latter are subsumed. Step **2.c** replaces $(\neg x_i \vee \cdots \vee \neg x_k \vee \phi, \alpha)$ by $(\phi, \alpha)$ since $(\neg x_i \vee \cdots \vee \neg x_k \vee \phi, \alpha)$ and $\{(x_i, 1), \cdots, (x_k, 1)\}$ implies $(\phi, \alpha)$ which subsumes $(\neg x_i \vee \cdots \vee \neg x_k \vee \phi, \alpha)$.

Now, assume that $B \cup \{(x_1, 1), \cdots, (x_n, 1)\} \vdash (a_1, \alpha)$ (Step **2.d** and **2.e**). Let $(\phi, \delta) \in B$, such that $\delta > \alpha$. Then, variables which are in $\phi$ should be added to $Par(A_1)$. Indeed, let $\phi = v_1 \vee \cdots \vee v_n$. Then, one can easily check that:

$\Pi(a_1 \mid x_1 \cdots x_n \neg v_1 \cdots \neg v_n) \neq \Pi(a_1 \mid x_1 \cdots x_n)$,
because $B \cup \{(x_1, 1), \cdots, (\neg v_1, 1), \cdots, (\neg v_n, 1)\}$ is inconsistent to a degree $\geq \delta$ from which $a_1$ can no longer be inferred.

**Example 7** *Let us consider again the base* $\Sigma = \{(su \vee \neg wi, \frac{2}{3}), (\neg wi \vee se, \frac{1}{3}), (wi \vee \neg se, \frac{1}{3}), (su \vee se, \frac{1}{3})\}$. *Then, using the above algorithm we get:*
$Par(SE) = \{WI, SU\}$,
$Par(WI) = \{SU\}$ *and* $Par(SU) = \emptyset$.

## 6 Computation of conditional possibility degrees

This subsection shows how to compute $\Pi(A_1 \mid Par(A_1))$ once $Par(A_1)$ is fixed.
Let $(x_1, \cdots, x_n)$ be an instance of $Par(A_1)$, and $a_1$ an instance of $A_1$. Recall that by definition:
$$\Pi(a_1 \mid x_1 x_2 \cdots x_n) = \frac{\Pi(a_1 x_1 \cdots x_n)}{\Pi(x_1 \cdots x_n)},$$
and that $\Pi(a_1 \mid x_1 x_2 \cdots x_n) = 1$ if $\Pi(x_1 \cdots x_n) = 0$.
The following proposition provides the computation of $\Pi(\phi)$ syntactically:

**Proposition 5** $\Pi(\phi) = 1 - Inc(\Sigma \cup \{(\phi, 1)\})$.
*(We recall that $\Sigma$ is assumed to be consistent).*

The proof is immediate since $\Pi(\phi) = 1 - N(\neg \phi)$, and that $N(\neg \phi) = Inc(\Sigma \cup \{(\phi, 1)\})$ (See [6]).
Therefore, to compute $\Pi(a_1 \mid x_1 \cdots x_n)$:

1. Add $\{(x_1, 1), \cdots, (x_n, 1)\}$ to $\Sigma$. Let $\Sigma'$ be the result of this step.

2. Compute $h = 1 - Inc(\Sigma')$ ($h$ represents $\Pi(x_1 \cdots x_n)$)

3. Add $\{(a_1, 1)\}$ to $\Sigma'$. Let $\Sigma''$ be the result of this step.

4. Compute $h' = 1 - Inc(\Sigma'')$ ($h'$ represents $\Pi(a_1 x_1 \cdots x_n)$). Then,
$$\Pi(a_1 \mid x_1 \cdots x_n) = \begin{cases} 1 & \text{if } h = 0 \\ \frac{h'}{h} & \text{otherwise.} \end{cases}$$

**Example 7** (continued)
*Let us illustrate the computation of* $\Pi(\neg se \mid wi \wedge su)$.
*We add the instance* $\{(wi, 1), (su, 1)\}$ *to* $\Sigma$ *(Step 1).*
*We get* $\Sigma' = \{(wi, 1), (su, 1), (su \vee \neg wi, \frac{2}{3}), (su \vee se, \frac{1}{3}), (\neg wi \vee se, \frac{1}{3}), (wi \vee \neg se, \frac{1}{3})\}$.
*We have* $Inc(\Sigma') = 0$. *Then,* $h = 1$ *(Step 2).*
*We now add* $(\neg se, 1)$ *to* $\Sigma'$ *(Step 3).*
*We get* $\Sigma'' = \{(\neg se, 1), (wi, 1), (su, 1), (su \vee \neg wi, \frac{2}{3}), (su \vee se, \frac{1}{3}), (\neg wi \vee se, \frac{1}{3}), (wi \vee \neg se, \frac{1}{3})\}$.
*We have* $Inc(\Sigma'') = \frac{1}{3}$. *Then,* $h' = \frac{2}{3}$.
*Hence,* $\Pi(\neg se \wedge wi \wedge su) = \frac{2}{3}$. *Lastly:* $\Pi(\neg se \mid wi \wedge su) = \frac{h'}{h} = \frac{2}{3}$. *With a similar way, we get the following conditional possibilities:*

$\Pi(SU)$

| su | 1 |
|---|---|
| $\neg su$ | $\frac{2}{3}$ |

$\Pi(WI|SU)$

|  | $\neg su$ | su |
|---|---|---|
| wi | $\frac{1}{2}$ | 1 |
| $\neg wi$ | 1 | 1 |

$\Pi(SE \mid WI, SU)$

|  | $\neg wi su$ | $\neg wi \neg su$ | $wi su$ | $wi \neg su$ |
|---|---|---|---|---|
| se | $\frac{2}{3}$ | 1 | 1 | 1 |
| $\neg se$ | 1 | 1 | $\frac{2}{3}$ | 1 |

*It can be checked that the possibility distribution associated with the constructed* $\Pi G$ *(using the chain rule) is the same as the one associated with the possibilistic base.*
Observe that we have chosen the ordering of the variables in the example arbitrarily. Clearly each application suggests orderings which are the more natural ones, or which leads to a simple structure. Note also that the computation of the weights could be also handled symbolically.

## 7 Conclusion

In the possibility theory framework, desires or knowledge can be equivalenty expressed in different formats. This paper has used two compact representations of possibility distribution: a *possibilistic knowledge base* and *possibilistic graph*.
Each of these representations have been shown in previous papers [3, 6] to be equivalent to a possibility distribution which rank-orders the possible worlds according to their level of plausibility. The framework may use a symbolic discrete linearly ordered scale, or can as well be interfaced with numerical settings by using the unit interval as a scale, using a different type of conditioning in each case [9].
This paper is on step further in establishing the relationship between different compact representation of possibility distribution. The results presented in this paper would enable us also to translate a possibilistic logic base easily into a kappa function graph [14] since there exists direct transformations [7] between possibility theory and Spohn's ordinal conditional functions [19].